\documentclass[runningheads]{llncs}

\usepackage[table,xcdraw]{xcolor}
\usepackage{colortbl}

\usepackage{booktabs} 
\usepackage{subfig}
\usepackage{graphicx}
\usepackage{multirow}

\usepackage{amsmath}

\usepackage{array}
    \newcolumntype{P}[1]{>{\arraybackslash}p{#1}}
    \newcolumntype{M}[1]{>{\centering\arraybackslash}m{#1}}

\usepackage{footnote}
\usepackage[bottom]{footmisc}
\makesavenoteenv{tabular}
\makesavenoteenv{table}

\usepackage{standalone} 
\usepackage{tikz}
\usetikzlibrary{calc}
\usepackage{physics}
\usepackage{amsmath}
\usepackage{mathdots}
\usepackage{yhmath}
\usepackage{cancel}
\usepackage{color}
\usepackage{siunitx}
\usepackage{array}
\usepackage{multirow}
\usepackage{amssymb}
\usepackage{tabularx}
\usepackage{extarrows}
\usepackage{booktabs}
\usetikzlibrary{fadings}
\usetikzlibrary{patterns}
\usetikzlibrary{shadows.blur}
\usetikzlibrary{shapes}

\begin{document}

\title{Meme Sentiment Analysis Enhanced with Multimodal Spatial Encoding and Face Embedding\thanks{This work was conducted with the financial support of the Science Foundation Ireland Centre for Research Training in Digitally-Enhanced Reality (d-real) under Grant No. 18/CRT/6224.}}

\titlerunning{Multimodal Spatial Encoding in Sentiment Classification of Memes.}

\author{Muzhaffar Hazman\inst{1}\orcidID{0000-0001-8262-2476}\and \\ 
Susan McKeever\inst{2}\orcidID{0000-0003-1766-2441} \and \\
Josephine Griffith\inst{1}\orcidID{0000-0002-1560-1867}
}

\authorrunning{M. Hazman, S. McKeever, J. Griffith.}

\institute{University of Galway, Galway, Ireland\\
\email{\{m.hazman1, josephine.griffith\}@universityofgalway.ie}
\and
Technological University Dublin, Dublin, Ireland \\
\email{susan.mckeever@TUDublin.ie}}
\maketitle              
\begin{abstract}

Internet memes are characterised by the interspersing of text amongst visual elements. State-of-the-art multimodal meme classifiers do not account for the relative positions of these elements across the two modalities, despite the latent meaning associated with where text and visual elements are placed. Against two meme sentiment classification datasets, we systematically show performance gains from incorporating the spatial position of visual objects, faces, and text clusters extracted from memes. In addition, we also present facial embedding as an impactful enhancement to image representation in a multimodal meme classifier. Finally, we show that incorporating this spatial information allows our fully automated approaches to outperform their corresponding baselines that rely on additional human validation of OCR-extracted text.

\end{abstract}

\section{Introduction}
The sentiment polarity classification task traditionally entailed analysing a piece of natural language text to classify its sentiment as negative, positive, or neutral. Sentiment analysis was initially performed on text. The growth of user-generated multimodal content (e.g., videos, image-caption pairs) has motivated the extension of affective computing techniques to input types beyond text \cite{poria_review}. Multimodal sentiment analysis poses the same questions as its text-only predecessor, but is extended to inputs comprising multiple modalities simultaneously. When faced with multimodal inputs, Poria et al. \cite{poria_review} describe unimodal encoders as crucial building blocks of multimodal systems, each encoder directly contributing to the resultant performance. Furthermore, the fusion of unimodal representations also plays a key role by providing ``surplus information'' to the classifier \cite{poria_review}.

\begin{figure}[t]

\centering
\includegraphics[width=0.8\textwidth]{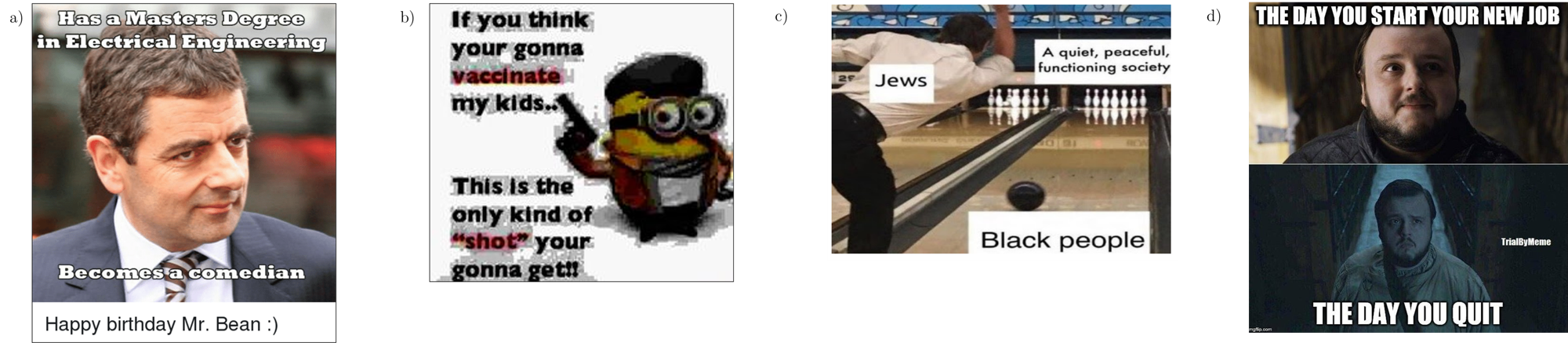}

\caption{Sample memes with a) \textit{Positive} sentiment\cite{memo2_report} and b) \textit{Negative} sentiment\cite{memo2_report}, c) hateful spatial analogies\cite{aomd}, and d) spatial segments\cite{memo1_report}.}
\label{fig:figure3}
\vspace{-10pt}
\end{figure}

Along with the advent of other multimodal formats of user-generated content, Internet memes (or simply ``memes'') have proliferated. Memes are commonly found in various online communities to communicate ideas, incite humour, and express emotions. Automated analysis of memes allows for: including memes in automated opinion mining processes \cite{poria_review}, taking action against meme-based hate speech \cite{aomd,hateful_report}, identifying disinformation campaigns \cite{meme_fakenews}, and investigating social and political cultures \cite{phomemes_exploiting}. This work contributes to the underlying problem of \textbf{sentiment polarity classification of a meme}: ``Given a meme in a visual format, comprising an image $I$ with embedded text $T$, classify the meme as having the overall sentiment of either \textit{Negative} (e.g., Fig. 1b), \textit{Positive} (e.g., Fig. 1a), or \textit{Neutral}''. 

Memes are challenging input in automated affective classification problems, as they typically exhibit very brief texts, references to popular culture, subtle intermodal semantic relations, and dependence on background context \cite{momenta,aomd,zhu,aomd}. Thus, solutions must consider the semantics of each, the textual and visual modalities, and their combinations \cite{hateful_report}. The breadth of this challenge spans various affective goals, including sentiment polarity \cite{memo2_report,memo1_report}, offensiveness \cite{hateful_report,memo2_report,memo1_report}, sarcasm \cite{memo2_report,memo1_report}, and motivational intent \cite{memo2_report,memo1_report}.

Recent work has shown that incorporating additional relevant information improves the performance of meme affective classifiers \cite{momenta}, amongst which is positional information of words within text and visual objects within an image \cite{aomd,zhu}. Unlike many other forms of multimodal content, the text within a meme is interspersed into its image, often either superimposed on the image or comprising a segment of the meme image, creating a shared visual medium. Meme authors intentionally position a grouping of words (``text clusters'') to convey meaning, such as implying hateful analogies \cite{aomd} (e.g., Fig. 1c); text clusters can be paired with image segments, with each pair signifying a different sentiment (e.g., Fig. 1d). Current approaches that use positional information in meme sentiment classification opt to omit intermodal positional relations, i.e. they consider the position of a word amongst text but not its position in relation to the meme image or vice versa. 

\begin{figure}[!htbp]

\centering
\includegraphics[width=0.7\textwidth]{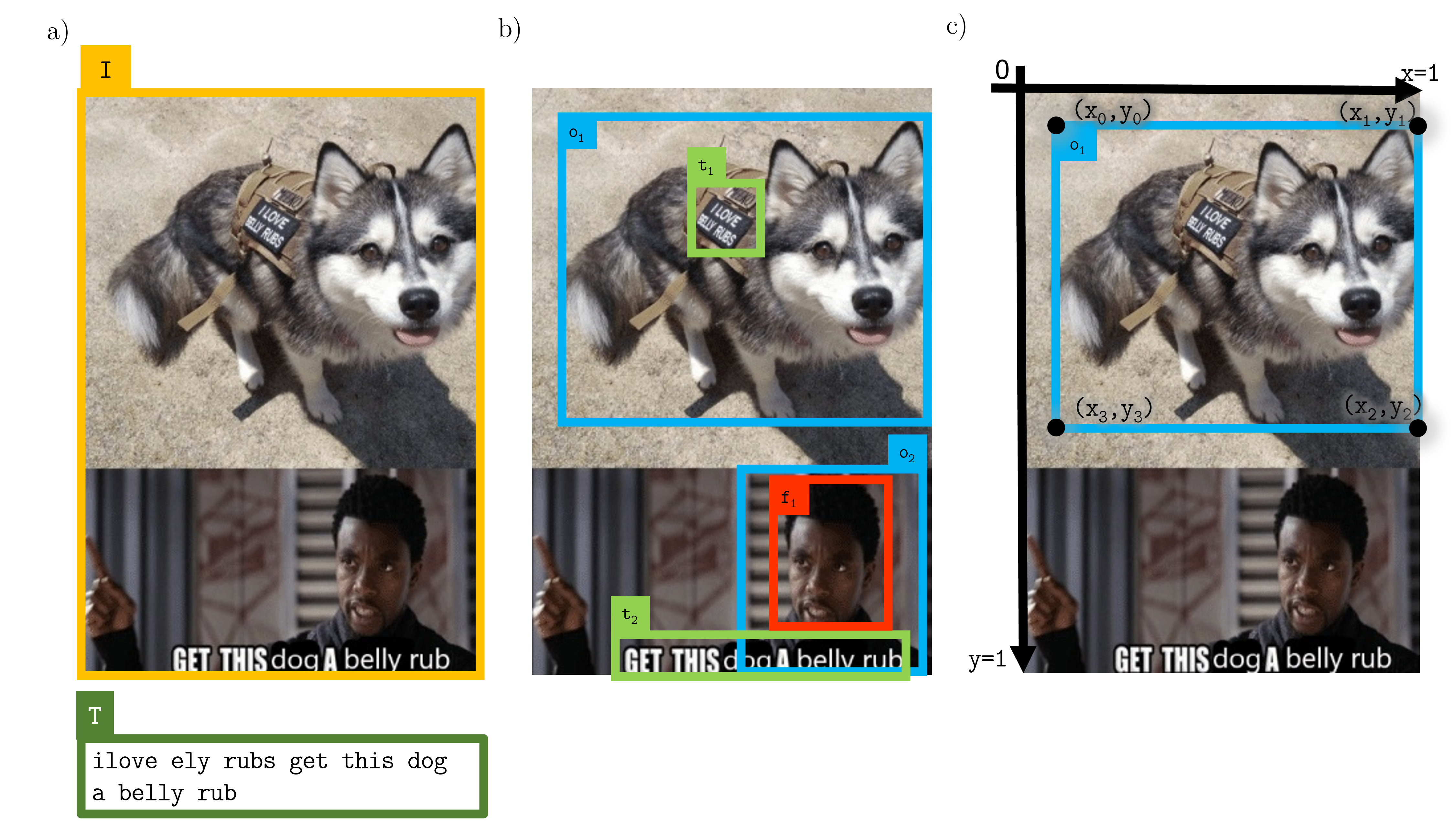}

\caption{Sample meme \cite{memo2_report} a) showing the image and text modalities, $I$ and $T$, as given in the dataset; b) bounding boxes generated for local features: text clusters ($t_{1}$ and $t_{2}$), objects ($o_{1}$ detected as ``Dog'' and $o_{2}$ as ``Person''), and faces ($f_{1}$); and c) the coordinate system used to generate the spatial encoding for each bounding box (e.g. the vertices of $o_1$, $p_{o_1}$).}

\end{figure}

 This work proposes injecting the spatial information of features from both modalities of a meme into a deep learning multimodal classifier to improve sentiment classification performance. Crucially, we account for the interspersing of visual objects and text clusters by representing the spatial position of each on a shared coordinate system (``spatial encoding''). We append the spatial encoding of visual objects (e.g. $o_1,o_2$ in Fig. 2b), faces (e.g. $f_1$ in Fig. 2b), and text clusters (e.g. $t_1,t_2$ in Fig. 2b) to their local representations prior to multimodal fusion and classification. The performance implication of spatial encodings and local representations are systematically evaluated on two benchmark datasets using the seven models described in Section 3.2. To the best of our knowledge, this work is the first to use shared coordinate spatial encoding and deep representation of faces to tackle the sentiment classification of memes.

\section{Related Works}

\subsection{Meme Affective Classifiers}

Memes are distinct from other multimodal user-generated content types in several key ways. First, the text and image of a meme share a common visual medium, unlike the more common image-caption pairs. Text in memes is often intentionally located amongst other visual content to create meaning \cite{aomd}.  Second, memes use short text pieces and few foreground visual objects, relying on intermodal relations to convey meaning. Kiela et al. \cite{hateful_report} show how harmless images and texts could be combined to create hateful memes. Furthermore, slight changes in either modality can change a hateful meme into a harmless one and vice versa. Therefore, meme classifiers must be able to learn subtle intermodal relationships with very limited input. 
\par
Architecturally, the current literature suggests that various affective classification tasks can be applied to memes without requiring entirely distinct approaches. Most apparently, Bucur et al.'s \cite{blue} winning submission of the Memotion 2022 Challenge \cite{memo2_report}, was trained to simultaneously classify sentiment polarity, offensiveness, sarcasm, humour, and motivational intent. Their findings suggest that meme classification architectures exhibit adaptability across different affective computing tasks. Furthermore, Pramanick et al. \cite{mhameme}, who reported the best-performing sentiment classification solution to the Memotion 1.0 dataset \cite{memo1_report}, showed that the same architecture outperforms all, or all but one, competing solution when individually trained on eight affect dimensions. 
\par
A typical approach to building a multimodal meme classifier is to generate unimodal representations of each modality before fusing these representations into a multimodal representation of the meme, such as in \cite{blue,momenta,aomd,mhameme}. Furthermore, the literature presents a wide range of deep learning representations used for each visual and textual modality \cite{hateful_report,memo2_report,memo1_report}, with no clear evidence that any of the options would consistently outperform all others.

\subsection{Positional Encoding} 
Positional encoding plays a central role in the Transformers architecture \cite{transformers} and has seen wide adoption in tackling various natural language tasks. It describes the position of tokens, such as a word in a sentence or a region in an image, within the input. However, since most multimodal meme classifiers employ unimodal encoders, the positions of text and visual elements are encoded separately. 
\par
To the best of our knowledge, a positional encoding that is shared between the text and image modalities on a common spatial coordinate system (a ``spatial encoding'') has not been applied to classifying meme sentiment. None of the architectures  reportedly used to learn meme sentiment classification in \cite{memo1_report} and \cite{memo2_report} did so using a positional information from a coordinate system shared between modalities. Further, we were not able to find a pre-trained multimodal Transformer that readily supports such a shared encoding.
\par
In this task, Pramanick et al. \cite{mhameme} showed performance gains by segmenting the text modality into text clusters but did not explicitly represent the spatial position of each cluster. To classify hateful memes, Zhu \cite{zhu} employed a patch detector to divide each meme into ``image regions''. They then appended each text token with a representation of its surrounding image patch. However, they did not present the performance gains solely attributable to this approach. Further, we posit that such a patch-based definition of position would not be suitable where multiple text clusters are placed within the same image patch (e.g., Fig. 1c) or where a patch consists only of text (e.g. Fig 1a). 
\par Shang et al. \cite{aomd} proposed a more general representation of spatial position by appending the spatial encoding of extracted visual objects and text clusters prior to input into an intermodal co-attentive pooling module based on a design from \cite{coattention}. They attributed their model's outperforming of other leading hateful meme classifiers to its ``awareness'' of offensive intermodal analogies: the purposeful superimposing of a text cluster near to a visual object is used to represent an offensive conceptual comparison. While their approach is predicated solely on offensive spatial analogies, we posit that this approach could capture a broader category of intermodal spatial relationships, including those captured by Pramanick et al.'s \cite{mhameme} and Zhu's \cite{zhu} approaches.

\subsection{Visual Feature Representations} 
While the image modality is commonly represented by passing the entire meme image through an image encoder \cite{memo2_report}, enhancing this representation with that of extracted visual objects has proven beneficial in classifying hateful memes \cite{momenta,aomd,zhu}. One such approach is to input the meme image into Google Cloud Vision API's Web Entity Detection to create a corresponding description or set of attributes in text format \cite{momenta,zhu}. Zhu \cite{zhu} also demonstrated further performance improvement with the inclusion of Race and Gender tags for each face using a pre-trained FairFace classifier. Pramanick et al. \cite{momenta} also showed improved performance by representing cropped images of visual objects and faces with VGG-19. Shang et al. \cite{aomd} also found that their multimodal classifiers perform best when global and local visual feature representations are available.
\par
The use of faces to convey sentiment is neither new nor unique to memes. Firstly, visual sentiment analysis \cite{sentribute} points to facial expressions as a valuable mid-level feature in classifying the sentiment conveyed by images from social networks. Second, facial expression emojis have been shown to be informative in supporting the sentiment classification of textual social media \cite{effectsentiment_emojis}. In memes, Zhu \cite{zhu} argues that expecting a global image encoding to sufficiently recognise facial features that are predictive of hatefulness is unreasonable given the size of current meme datasets. Although we agree with Zhu's argument, we posit that their approach omits other information conveyed by faces that may indicate a meme's sentiment, such as emotion, expression, and identity.

\section{Methodology}
In this work, we evaluate the performance of seven novel multimodal classifier models. These models are separately trained on two competition datasets, Memotion 1.0 \cite{memo1_report} and Memotion 2.0, \cite{memo2_report}, to classify the sentiment polarity of memes. We first designed and evaluated a multimodal deep learning model to establish baseline performance. This model is then repeatedly augmented to answer our research questions. Augmentations include incorporating spatial information of faces, visual objects, and text clusters and are described for each model in Table 3. Evaluation is conducted based on the differences in macro-averaged and weighted-averaged F1 scores -- metrics prescribed by the authors of the datasets \cite{memo2_report,memo1_report} -- between pairs of models that respectively include and exclude each augmentation. This section presents details of the datasets and models used.

\subsection{Dataset \& Feature Extraction} 
\begin{table}[!t]
\centering
\caption{Samples per dataset.\label{tab_data}}

\begin{tabular}{lcccccc}
\toprule
  &  \multicolumn{3}{c}{\textbf{Memo1}} & \multicolumn{3}{c}{\textbf{Memo2}} \\
\cmidrule(lr){2-4} \cmidrule(lr){5-7}
\multirow{-2}{*}{\textbf{Dataset}} & \textbf{Train} & \textbf{Val} & \textbf{Test} & \textbf{Train} & \textbf{Val} & \textbf{Test}\\
\midrule
\texttt{\textbf{Original}} & & & & & &\\

 \hspace{5pt}Positive & 4,156 & -- & 1,099 & 1517 & 325 & 78 \\
 \hspace{5pt}Neutral & 2,204 & -- & 584 & 4510 & 975 & 971 \\
 \hspace{5pt}Negative & 631 & -- & 172 & 973 & 200 & 451\\
 \hspace{5pt}\textbf{Total }& \textbf{6,991} & -- & \textbf{1,855} & \textbf{7,000} & \textbf{1,500} & \textbf{1,500} \\

\arrayrulecolor{lightgray}
\midrule

\multicolumn{2}{l}{\textbf{\texttt{Filtered \& }} } & & & & & \\
\multicolumn{2}{l}{\textbf{\texttt{Filtered-OCR}} } & & & & & \\
\hspace{5pt}Positive & 3,450 & 609 & 1,067 & 1,453 & 192 & 76 \\
\hspace{5pt}Neutral & 1,837 & 324 & 572 & 4,363 & 951 & 939\\
\hspace{5pt}Negative & 518 & 92 & 169 & 941 & 317 & 442\\
\hspace{5pt}\textbf{Total} & \textbf{5,805} & \textbf{1,025} & \textbf{1,808} & \textbf{6,757} & \textbf{1,460} & \textbf{1,457}\\
\arrayrulecolor{black}
\bottomrule
\end{tabular}
\end{table}
\par This work utilises datasets presented in the SemEval 2019 Memotion 1.0 \cite{memo1_report} (``\textbf{Memo1}'') and AAAI 2022 Memotion 2.0 \cite{memo2_report} challenges (``\textbf{Memo2}''). Both are collections of user-generated memes labelled with one of three exclusive sentiment classes. The authors of the datasets extracted text from each meme with an automated OCR tool and then manually corrected any erroneous text extraction. For our experiments, the samples from Memo1 and Memo2 are kept separate. Without filtering or pre-processing, these samples comprise our \texttt{\textbf{Original}} datasets that we use to compare our \texttt{Baseline} model to leading solutions.

\begin{figure}[!b]
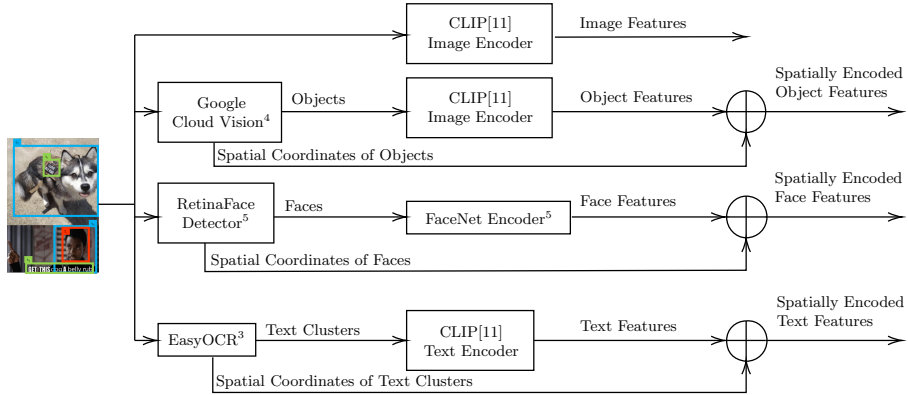


\centering
\includestandalone[width=0.99\textwidth]{diag_feature}

\caption{Localisation and representation process applied to each meme to extract its Image, Object, Face and Text features.}

\vspace{-10pt}

\end{figure}
\footnotetext[3]{https://github.com/JaidedAI/EasyOCR; \texttt{paragraph} option set to true.}
\footnotetext[4]{https://cloud.google.com/vision/docs/object-localizer}
\footnotetext[5]{Using DeepFace wrapper from https://github.com/serengil/deepface}

\par For each meme in these datasets, we localised, extracted, and represented its text clusters, faces, and visual objects using the tools listed in Fig. 3. The maximum counts of text clusters, visual objects, and faces are set to 18, 10, and 5, respectively, with padding used for memes with fewer. Padding for text clusters is defined by passing an empty string into the CLIP text encoder, while that for visual objects is the CLIP encoding of a blank image, and zero--padding is used for faces.

Since this work applies to memes that contain identifiable visual objects and text clusters, we removed meme samples that do not meet these criteria to make up the \texttt{\textbf{Filtered}} datasets. This filtering is performed on all subsets of Memo1 and Memo2. As Memo1 did not contain a designated validation set, we defined one by splitting the training set -- as reported by the authors of the Memo1 dataset and used in submissions to their competition \cite{memo1_report} -- with a random 85:15 sampling, weighted by the sentiment class, to maintain the target distribution. We maintained the train-validation-test splits defined for Memo2 \cite{memo2_report}. Meme samples with identifiable visual objects but no detected faces are given face feature representation made up entirely of padding.

Finally, the \textbf{\texttt{Filtered-OCR}} datasets replace the text of each meme in \texttt{Filtered} with that returned in our feature extraction OCR step. Unlike in \cite{memo2_report,mhameme,memo1_report}, we excluded any additional human validation during the OCR extraction process. All models are trained, validated, and tested on the resultant \texttt{Filtered-OCR} datasets. The counts of memes in each dataset and sentiment labels are shown in Table \ref{tab_data}.

\subsection{Models}

\begin{figure}[!t]
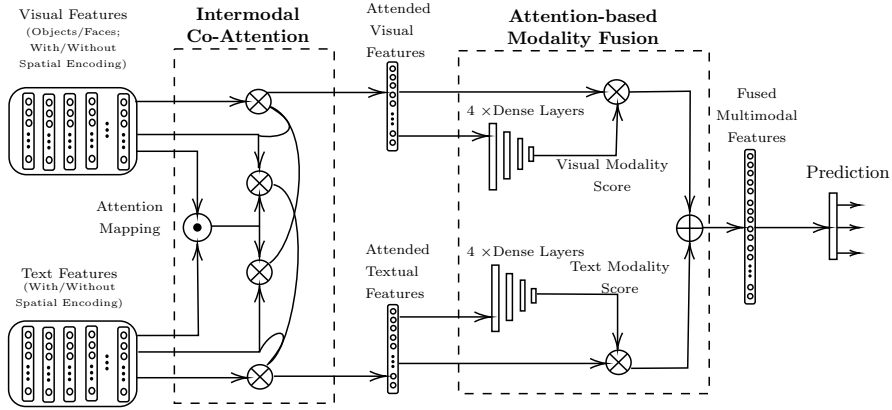


\centering
\includestandalone[width=\textwidth]{diag_arch}

\caption{Architecture of our \texttt{Obj-NoSpatial}, \texttt{Obj-Spatial}, \texttt{Face-NoSpatial}, and \texttt{Face-Spatial} models. The Image features used in \texttt{Img-Obj-Spatial} and \texttt{Img-Face-Spatial} models bypasses the Intermodal Co-Attention module and requires the Attention-Based Modality Fusion module to be expanded with another set of dense layers. This work's \texttt{Baseline} model does not include the Intermodal Co-Attention module. Sources: Intermodal Co-Attention \cite{coattention,aomd}; Attention-based Modality Fusion \cite{gu_fusion,mhameme} \label{diag_arch}}

\vspace{-10pt}
\end{figure}

\begin{table}[!b] \centering \caption{Goals of each experimental model. \label{tab_models}}
\resizebox{0.9\linewidth}{!}{%
\begin{tabular}{p{0.25\columnwidth} p{0.19\columnwidth}  p{0.55\columnwidth}}
\toprule
\textbf{Model} & \textbf{Dataset} & \textbf{Goal}  \\
\midrule

\arrayrulecolor{lightgray}
 \global\setlength{\aboverulesep}{0pt}
 \global\setlength{\belowrulesep}{0pt}
\texttt{\textbf{Baseline}} & \texttt{Original} &  Benchmarks our chosen modality encodings and fusion mechanism against leading solutions.\\
\cmidrule{2-3}
 & \texttt{Filtered} & Establishes baseline performance on samples with detectable text clusters and visual objects. \\
\cmidrule{2-3}
 & \texttt{Filtered-OCR} & Measures the impact of replacing human-curated text replaced with text clusters returned by automated OCR. Also, establishes a baseline for our fully automated approaches.\\
\midrule
\texttt{\textbf{Obj-NoSpatial}} & \texttt{Filtered-OCR} & Measures the performance of representing the image modality using only CLIP-encoded localised visual objects without spatial encodings. \\
\midrule
\texttt{\textbf{Obj-Spatial}} & \texttt{Filtered-OCR} & Measure the performance impact of including spatial encodings of objects and text clusters.\\
\midrule
\texttt{\textbf{Img-Obj-Spatial}} & \texttt{Filtered-OCR} & Maximises available visual information by augmenting image input with objects and text clusters and respective spatial encodings.\\
\midrule
\texttt{\textbf{Face-NoSpatial}} & \texttt{Filtered-OCR} & Measures the performance of representing the  image modality using only embeddings of localised faces without spatial encodings.\\
\midrule
\texttt{\textbf{Face-Spatial}} & \texttt{Filtered-OCR} & Measure the performance impact of including spatial encodings of faces and text clusters.\\
\midrule
\texttt{\textbf{Img-Face-Spatial}} & \texttt{Filtered-OCR} & Augments image input with faces and text clusters and respective spatial encodings.\\

\arrayrulecolor{black}
\bottomrule

\end{tabular}
\global\setlength{\aboverulesep}{0.605mm }
\global\setlength{\belowrulesep}{0.984mm}

}
\end{table}

\par
 This section describes the architectural characteristics of our models as listed in Table \ref{tab_models} and illustrated in Fig. \ref{diag_arch}. Each was built using PyTorch and trained with a triangular cyclical learning rate schedule ranging between $1\mathrm{e}{-4}$ and $1\mathrm{e}{-3}$ with a step size of 52 mini-batches of 512 samples. During training, validation performance was monitored for overfitting or until each model was trained for 100 epochs. Training is carried out using AdamW optimiser with weight decay of $5e-1$, betas of 0.1 and 0.25 to minimise negative log-likelihood loss with class weights inversely proportional to its sample count in the training dataset. All non-pretrained weights are initialised with a zero-mean Gaussian distribution with standard deviation 0.02 , while pretrained weights are not fine-tuned. The same hyperparameter settings are maintained across all models as they are separately trained on the datasets.

 \par Leading meme sentiment classifiers use a variety of architectures with little indication of which is most optimal. For our \textbf{\texttt{Baseline}} model, we drew inspiration from the typical overall approach used in leading solutions to the Memotion 2.0 Challenge \cite{memo2_report}: each modality is represented using a pretrained encoder. Then, these representations are fused, often with a multimodal attention mechanism, and finally passed to a fully connected layer. 
 \par To encode the meme image and text (see $I$ and $T$ in Fig. 2a) in our \textbf{\texttt{Baseline}} model, we opted to use the pretrained image and text encoders of CLIP \cite{clip}, respectively, which has shown comparable performance to other multimodal approaches \cite{momenta}. In addition, CLIP image encodings have been shown to outperform various other image encoders in the zero-shot classification of hateful memes \cite{clip} and are used by the winning solution of the Memo2 challenge \cite{blue}. We chose the ViT--B/16 variant of CLIP while Pramanick et al. \cite{momenta} and Bucur et al. \cite{blue} did not report their chosen variant. 
 \par Since attentive fusion has been shown to perform well on several meme problems \cite{mhameme}, we included one in our models. Our \texttt{Baseline} model fuses the CLIP representations of the meme image and text using Gu's \cite{gu_fusion} attentive modality fusion mechanism, as used in \cite{momenta}. We defined the sizes of the four dense layers as 256, 64, 8, and 1, which produces an attention score for each modality. The attention-weighted representation of each modality is concatenated and passed into a GeLU-activated dense layer followed by a log-softmax activation to output predicted logits of each sentiment class.

\par This model is trained on the \texttt{Original} dataset to allow performance comparisons with previously published works. We then evaluated this model on the \texttt{Filtered} and \texttt{Filtered-OCR} datasets. In the latter, the content of all text clusters $t_n$ is concatenated and entered into the text encoder. The difference in the performance of this model on these two datasets allows us to measure the performance impact resulting from our OCR-based text extraction output relative to the human-curated approach used by the authors of the datasets \cite{memo2_report,memo1_report}.
\par
The \textbf{\texttt{Obj-NoSpatial}} and \textbf{\texttt{Face-NoSpatial}} models remove the meme image and text, $I$ and $T$ per \texttt{Baseline}. As inputs, the former takes CLIP-encoded visual objects, $o_1,o_2,...,o_j$, and text clusters extracted from a meme, $t_1,t_2,...,t_i$. Instead of objects, the \texttt{Face-NoSpatial} model takes the FaceNet representation of faces, $f_1,f_2,...,f_k$. Then, the $j$ visual objects or $k$ face representations are passed through co-attentive weighted pooling against $i$ text  clusters as used in \cite{aomd} but without spatial encodings. This step allows the models to learn attention maps between each object/face and each text cluster; producing a one-dimensional vector representing each modality. This representation replaces that of the image modality as input into the attentive fusion mechanism described for the \texttt{Baseline} model. 
\par
The \textbf{\texttt{Obj-Spatial}} and \textbf{\texttt{Face-Spatial}} models introduce the spatial encodings of each text cluster, $p_{t_i}$, as well as for visual objects, $p_{o_j}$, and faces, $p_{f_k}$, respectively. We augment the co-attentive pooling module in \texttt{Obj-NoSpatial} and \texttt{Face-NoSpatial} into the co-attentive analogy alignment module proposed in \cite{aomd}. This is performed by appending each object's and cluster/face's representation vector with its spatial encoding. The padding for spatial encodings is defined as zeros for all coordinates.
\par
The \textbf{\texttt{Img-Obj-Spatial}} and \textbf{\texttt{Img-Face-Spatial}} models each combine the CLIP representation of the meme image, $I$, into \texttt{Obj-Spatial} and \texttt{Face-Spatial}, respectively. Since these models make use of three representations per meme -- image, text clusters and objects/faces -- we extend Gu's \cite{gu_fusion} fusion mechanism to accommodate three inputs by introducing a third set of dense layers.

\section{Results}
Evaluating the \texttt{Baseline} model on the \texttt{Original} datasets places it within the top six highest performing solutions on each respective dataset; see Tables \ref{tab_benchmark_memo1} and \ref{tab_benchmark_memo2}.

\begin{table}[!t]

\begin{minipage}[t]{.45\linewidth}
\centering

\caption{Performance of our \texttt{Baseline} model against leading solutions on the Memo1 dataset. Sources: \cite{memo1_report,mhameme}. \label{tab_benchmark_memo1}}
\begin{tabular}{lc}
\toprule

\textbf{Solution} & \textbf{Macro-F1} \\
\midrule
MHA-Meme\footnote[6]{Not a competition submission; results based on subset of the original dataset} \cite{mhameme} & \textbf{0.3762} \\
Vkeswani IITK & 0.3547 \\
Our \texttt{Baseline} & 0.3546 \\
Guoym & 0.3520 \\
Aihaihara & 0.3502 \\
Sourya Diptadas & 0.3476 \\
Irina Bejan & 0.3469 \\
\bottomrule
\end{tabular}

\end{minipage}
\hfill
\begin{minipage}[t]{.45\linewidth}
\centering

\caption{Performance of our \texttt{Baseline} model against leading solutions on the Memo2 dataset. Source: \cite{memo2_report}. \label{tab_benchmark_memo2}}
\begin{tabular}{lc}
 \toprule
 
 \textbf{Solution} & \textbf{Weighted-F1} \\
 \midrule
\textsc{Blue} & \textbf{0.5318} \\
\textsc{Browallia} & 0.5255 \\
Yeti & 0.5088 \\
Little Flower & 0.5081 \\
Greeny & 0.5037 \\
Our \texttt{Baseline} & 0.5035 \\
Amazon PARS & 0.5025 \\
\bottomrule
\end{tabular}

\end{minipage} 

\end{table}

\begin{table}[!b]
\centering
\caption{Weighted F1 (F1-W) and Macro F1 (F1-M) for the \texttt{Baseline} model on all datasets. \label{tab_results_baseline}}

\begin{tabular}{l@{\hskip 10pt}cc@{\hskip 10pt}cc}
\toprule
\multirow{2}{*}{\textbf{Dataset}}  &
\multicolumn{2}{c}{\textbf{Memo1}} &
\multicolumn{2}{c}{\textbf{Memo2}} \\
\cmidrule(lr){2-3} 
\cmidrule(lr){4-5}
&
\textbf{F1-W} &
\textbf{F1-M} &

\textbf{F1-W} &
\textbf{F1-M}  \\

\midrule


\texttt{Original}&

\textbf{0.481} &
\textbf{0.355} &

\textbf{0.504} &
\textbf{0.325} \\

\texttt{Filtered}&

0.475&
0.327 &

0.503 &
0.314 \\ 
\texttt{Filtered-OCR}&

0.462 &
0.326 &
0.439 &
0.283  \\ 
\bottomrule
 \end{tabular} 
\end{table}

The performance of the \texttt{Baseline} model on the \texttt{Original, Filtered} and \texttt{Filtered-OCR} datasets are shown in Table \ref{tab_results_baseline}. The lower performance of the model on the \texttt{Filtered} dataset than on the \texttt{Original} dataset likely stems from the removal of samples that contain only text on an object-less background. Classifying such samples is similar to discerning the sentiment of unimodal text inputs and is beyond the scope of this work. We attribute the performance decrease of the \texttt{Baseline} model on the \texttt{Filtered-OCR} vs. \texttt{Filtered} datasets to the lower quality of the text extracted with our automated OCR process relative to human-curated text. Despite this, our spatially aware models are able to overcome this performance penalty. The model that performs best on each dataset -- as seen in Table \ref{tab_results} -- constitutes \textbf{fully automated approaches} that outperform their respective \texttt{Baseline} models trained on the human-curated text from the \texttt{Filtered} datasets. By removing the need for manual intervention, fully automated models improve the feasibility of conducting sentiment classification of memes at scale, and reduce the effort necessary for creating future meme datasets.

\begin{table}[!t]
\centering
\caption{Weighted F1 (F1-W) and Macro F1 (F1-M) for all models on the Memo1 and Memo2 \texttt{Filtered-OCR} datasets. \textbf{Rel.} indicates relative performance to model stated in the \textbf{Comparison} column on each given dataset. \label{tab_results}}
\resizebox{\linewidth}{!}{%
\begin{tabular}{llcccccc}
\toprule
\multirow{2}{*}{\textbf{Model}}  &
\multirow{2}{*}{\textbf{Comparison}} &
\multicolumn{3}{c}{\textbf{Memo1}} &
\multicolumn{3}{c}{\textbf{Memo2}} \\
\cmidrule(lr){3-5} 
\cmidrule(lr){6-8}

& &
\textbf{F1-W} &
\textbf{F1-M} &
\textbf{Rel.} &
\textbf{F1-W} &
\textbf{F1-M} &
\textbf{Rel.}  \\

\midrule
\arrayrulecolor{lightgray}
\global\setlength{\aboverulesep}{0pt}
\global\setlength{\belowrulesep}{0pt}
\textbf{\texttt{Baseline}} &
- &
0.462 &
0.326 &
\cellcolor[HTML]{dddddd}\textbf{-} &
0.439 &
0.283 &

\cellcolor[HTML]{dddddd}\textbf{-} \\ 

\arrayrulecolor{lightgray}%
\midrule

\hspace{5pt}\textbf{\texttt{Obj-NoSpatial}} &
vs. \texttt{Baseline}  &
0.452 &
0.307 &
\cellcolor[HTML]{FF9999}\textbf{↓} &
0.427 &
0.271 &

\cellcolor[HTML]{FF9999}\textbf{↓} \\
\midrule

\hspace{5pt}\textbf{\texttt{Obj-Spatial}}&
vs. \texttt{Obj-NoSpatial} &
0.481 &
0.317 &
\cellcolor[HTML]{99FF99}\textbf{↑} &
0.482 &
0.305 &

\cellcolor[HTML]{99FF99}\textbf{↑}  \\ 

\midrule

\hspace{5pt}\textbf{\texttt{Img-Obj-Spatial}} &
vs. \texttt{Obj-Spatial}  &
\textbf{0.489} &
0.336 &
\cellcolor[HTML]{99FF99}\textbf{↑} &
0.499 &
0.300 &

\cellcolor[HTML]{FFFF99}\textbf{↑↓}\\
\midrule

\hspace{5pt}\textbf{\texttt{Face-NoSpatial}}&
vs. \texttt{Baseline}

&
&
&
\cellcolor[HTML]{99FF99}\textbf{↑} &
&
&

\cellcolor[HTML]{99FF99}\textbf{↑} \\ 

&
vs. \texttt{Obj-NoSpatial} &
\multirow{-2}{*}{0.476} &
\multirow{-2}{*}{0.340} &
\cellcolor[HTML]{99FF99}\textbf{↑}&
\multirow{-2}{*}{0.471} &
\multirow{-2}{*}{0.298} &
\cellcolor[HTML]{99FF99}\textbf{↑}  \\ 
\midrule

\hspace{5pt}\textbf{\texttt{Face-Spatial} }
&
vs. \texttt{Face-NoSpatial} 
&
&
&
\cellcolor[HTML]{99FF99}\textbf{↑} &
&
&

\cellcolor[HTML]{99FF99}\textbf{↑} \\ 

&
vs. \texttt{Obj-Spatial}  &
\multirow{-2}{*}{0.485} &
\multirow{-2}{*}{\textbf{0.341}} &
\cellcolor[HTML]{99FF99}\textbf{↑}&
\multirow{-2}{*}{0.496} &
\multirow{-2}{*}{0.310} &
\cellcolor[HTML]{99FF99}\textbf{↑} \\ 
\midrule

\hspace{5pt}\textbf{\texttt{Img-Face-Spatial}}&
vs. \texttt{Face-Spatial}
&
&
&
\cellcolor[HTML]{FF9999}\textbf{↓} &
&
&

\cellcolor[HTML]{99FF99}\textbf{↑}\\ 

&
vs. \texttt{Img-Obj-Spatial}  &
\multirow{-2}{*}{0.473} &
\multirow{-2}{*}{0.332} &
\cellcolor[HTML]{FF9999}\textbf{↓} &
\multirow{-2}{*}{\textbf{0.509}} &
\multirow{-2}{*}{\textbf{0.314
}} &
\cellcolor[HTML]{99FF99}\textbf{↑}
\\ 
\arrayrulecolor{black}
\bottomrule
 \end{tabular} }
\end{table}

\par
The results show that spatial encoding improves performance. \texttt{Obj-Spatial} and \texttt{Face-Spatial} each outperforms \texttt{Obj-NoSpatial} and \texttt{Face-NoSpatial} respectively. These results point to intermodal spatial information being informative for the problem task and not sufficiently represented by the CLIP encodings of the whole meme image. This finding holds significance to applying deep learning solutions on memes in particular, as the text modality is incorporated and interspersed within the image. Although the importance of token positions in leading solution architectures has been well established, the lack of a shared visual medium for image and text modalities in many other vision-language tasks has resulted in leading multimodal architectures with separate positional representations for each modality. Based on our results, we argue that spatial encodings should also be considered for other vision-language tasks where visual objects and text share a common visual medium.
\par
The performance benefit of representing the image modality with localised visual feature representations depends on whether the features are defined as objects or faces. CLIP-encoded object representation performs worse than \texttt{Baseline}. This results from a reduction in the visual information available to the image encoder. However, \texttt{Face-NoSpatial}, which uses FaceNet embeddings to represent faces, outperforms both \texttt{Obj-NoSpatial} and \texttt{Baseline} while also suffering from the same, if not greater, reduction in available visual information. Furthermore, \texttt{Obj-Spatial} showed mixed results against \texttt{Baseline}, while \texttt{Face-Spatial} outperforms \texttt{Baseline} in both datasets. Notably, faces are not entirely excluded from models based on visual objects, as many meme samples had ``Person'' as a detected object. Thus, we believe that the performance difference between the two approaches arises from the more fine-grained facial embedding provided by FaceNet and the inherent exclusion of non-face visual objects that emphasises the contribution of faces to the sentiment of a meme.
\par
We found that augmenting the meme image with local representations of either objects or faces and their spatial encodings consistently outperforms models that rely on the image alone. However, choosing between CLIP-encoded objects versus FaceNet-encoded faces as augmentations to the meme image proved inconsistent and dependent on the dataset. Although \texttt{Img-Obj-Spatial} and \texttt{Img-Face-Spatial} perform the best in the Memo1 and Memo2 datasets, respectively, their performance relative to \texttt{Obj-Spatial} and \texttt{Face-Spatial} appears to depend on the dataset. Drops in performance here may stem from redundant intermodal information (e.g. between global image and objects-based representations). Unlike in \cite{mhameme}, we did not employ any form of learned cross-modal filtering.

\section{Conclusions}
In this work, we addressed spatial encoding and facial embedding in classifying sentiment polarity of internet memes. We developed seven novel architectures, and evaluated each on two challenge datasets. For both datasets, our proposed baseline multimodal classifier ranked within the top six of leading state-of-the-art solutions on both datasets. While we found that representing the image modality with visual objects alone does not consistently offer performance benefits, a face-based representation does. Furthermore, the incorporation of spatial information of these visual features grants performance improvements over both image-only and faces-/objects-only approaches. For each of the Memotion datasets, our top performing solution comprises augmenting the image modality with spatially encoded visual features and text clusters. We propose these solutions as fully automated competitive alternatives to current state-of-the-art solutions that rely on manual validation of OCR-based text extraction.

\bibliographystyle{splncs04}
\bibliography{biblio}

\end{document}